\documentclass{article}

\usepackage{microtype}
\usepackage{graphicx}
\usepackage{subfigure}
\usepackage{booktabs}

\usepackage{hyperref}

\usepackage[accepted]{icml2025}
\makeatletter
\renewcommand{\Notice@String}{Preprint.}
\makeatother
\hypersetup{pdfsubject={Preprint}}

\usepackage{amsmath}
\usepackage{amssymb}
\usepackage{mathtools}
\usepackage{amsthm}
\DeclareMathOperator*{\argmax}{arg\,max}

\usepackage[capitalize,noabbrev]{cleveref}

\theoremstyle{plain}

\theoremstyle{definition}

\theoremstyle{remark}

\icmltitlerunning{Representational Difference Distillation}

\begin{document}

\twocolumn[
\icmltitle{Representational Difference Distillation}

\icmlsetsymbol{equal}{*}

\begin{icmlauthorlist}
\icmlauthor{Abdulkarim Mugisha}{equal,caltech}
\icmlauthor{Divin Irakiza}{equal,caltech}
\end{icmlauthorlist}

\icmlaffiliation{caltech}{Department of Computing + Mathematical Sciences, California Institute of Technology, Pasadena, US}
\icmlcorrespondingauthor{Abdulkarim Mugisha}{amugisha@caltech.edu}
\icmlcorrespondingauthor{Divin Irakiza}{dirakiza@caltech.edu}

\icmlkeywords{Machine Learning, Knowledge Distillation, Representational Difference, Metric Learning, Curriculum Learning.}

\vskip 0.3in
]

\printAffiliationsAndNotice{\icmlEqualContribution}

\begin{abstract}
Knowledge distillation transfers representational knowledge from a large teacher network to a compact student. Existing relational and contrastive distillation methods treat all samples uniformly, spending equal gradient on well-matched and poorly-matched regions of the representation space alike. We propose Representational Difference Distillation (RDD), which diagnoses where teacher and student representations structurally disagree and concentrates the distillation signal on those regions. RDD uses two asymmetric affinity matrices that capture opposing directions of misalignment: one identifies samples the student incorrectly considers similar, the other identifies teacher relationships the student has yet to learn. These affinities determine which sample pairs the distillation loss aligns and how strongly it corrects each misaligned sample. On CIFAR-100 with resnet32x4$\to$resnet8x4, RDD reaches 75.58\% test accuracy in our runs, on par with CRD (75.45\%) and above KD and RKD. We additionally explore curriculum-based and sampling-based variants, finding that weighting the loss is more effective than filtering the data.
\end{abstract}

\section{Introduction}
\label{submission}
Knowledge distillation transfers learned representations from a high-capacity teacher network to a compact student network, enabling deployment under computational constraints without severe performance degradation. The seminal formulation by \citet{hinton2015distillingknowledgeneuralnetwork} minimizes the KL divergence between teacher and student output distributions, effectively transferring  knowledge encoded in soft class probabilities. While simple and effective, this objective treats output dimensions independently and ignores the rich structural relationships within the teacher's representation space.
Subsequent work has addressed this limitation in two directions. Relational approaches such as RKD \cite{park2019relationalknowledgedistillation} transfer pairwise distance or similarity structure between samples, encouraging the student to preserve the teacher's inter-sample geometry. Contrastive approaches such as CRD \cite{tian2020contrastive} contrast paired teacher and student features against negative samples drawn from a memory bank, maximizing a bound on their mutual information, while RRD \cite{giakoumoglou2025relationalrepresentationdistillation} relaxes this objective into matching similarity distributions. These methods generally outperform pointwise matching objectives, suggesting that structural knowledge is important for effective distillation.

\quad However, existing relational and contrastive methods share a common limitation: they treat all samples uniformly. Every pair receives equal attention in the loss, regardless of whether the student has already captured that particular structural relationship or is failing to do so. This is wasteful since a well-trained student may already match the teacher's structure for the majority of samples, while struggling in specific regions where the representations fundamentally disagree. Allocating an equal gradient budget to the mastered and confused regions dilutes the learning signal.
We draw inspiration from Representational Difference Explanations (RDX), a recently proposed method that compares two representations by computing rank-normalized pairwise distances and applying a locally-biased difference function that amplifies disagreements between nearby samples while suppressing those between distant ones \cite{kondapaneni2025representationaldifferenceexplanations}. RDX provides a signal that existing distillation methods do not use: a per-sample measure of where the teacher and student structurally disagree. 

\begin{figure*}[t]
\vskip 0.1in
\begin{center}
\centerline{\includegraphics[width=\textwidth]{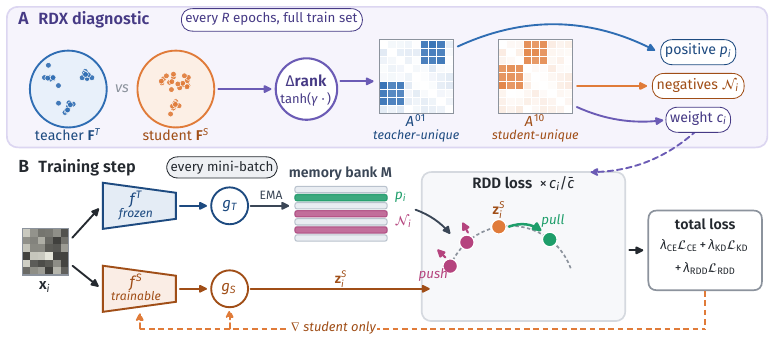}}
\caption{\textbf{Overview of Representational Difference Distillation.}
\textbf{(A)} Every $R$ epochs, RDX compares teacher and student embeddings of the training set through neighbor-rank differences, giving teacher-unique ($A^{01}$) and student-unique ($A^{10}$) affinities. These yield, for each sample $i$, a positive $p_i$, a pool of negatives $\mathcal{N}_i$, and a confusion weight $c_i$.
\textbf{(B)} At each step, the student embedding $\mathbf{z}^S_i$ is pulled toward the memory-bank entry of $p_i$ and pushed away from entries in $\mathcal{N}_i$, with the loss weighted by $c_i$. Only the student and $g_S$ receive gradients.}
\label{fig:rdd_overview}
\end{center}
\vskip -0.2in
\end{figure*}

\quad We propose Representational Difference Distillation (RDD), see Figure \ref{fig:rdd_overview}, a knowledge distillation method that uses RDX to target the regions of representational space where teacher and student disagree. RDD periodically computes the RDX locally-biased rank difference over the training set, producing a per-sample confusion score that quantifies structural misalignment. This score reweights a multi-negative contrastive objective, focusing gradients on samples whose neighborhood structure deviates most from the teacher while down-weighting already aligned regions. The contrastive loss is implemented with a momentum-updated memory bank of projected teacher features \cite{wu2018unsupervised,tian2020contrastive}, enabling thousands of negatives per anchor without extra forward passes.

\section{Related Work}

\subsection{Knowledge Distillation}
Model compression by training a small network to mimic a large one was introduced by \citet{bucila2006}, and knowledge distillation was popularized by \citet{hinton2015distillingknowledgeneuralnetwork}, who proposed matching softened output distributions via KL divergence with a temperature parameter. This formulation captures inter-class relationships encoded in the teacher's soft predictions, providing richer supervision than hard labels alone. FitNets \cite{romero2015fitnetshintsdeepnets} extended distillation to intermediate representations by regressing student feature maps toward teacher feature maps through learned projections. Attention Transfer \cite{zagoruyko2017payingattentionattentionimproving} instead matches spatial attention maps between teacher and student, focusing on where each network attends rather than the raw activations. These methods operate on individual samples independently, transferring pointwise activation patterns without explicitly considering relationships between samples.

\subsection{Relational and Contrastive Distillation}

Rather than matching pointwise features, relational methods transfer structural relationships between samples. RKD \cite{park2019relationalknowledgedistillation} penalizes differences in pairwise distances and angles, SP \cite{tung2019similarity} matches batch-wise similarity matrices, CC \cite{peng2019correlation} matches inter-sample correlations, and PKT \cite{passalis2018pkt} matches probability distributions over pairwise similarities. Contrastive methods push this further: CRD \cite{tian2020contrastive} maximizes a lower bound on mutual information between teacher and student representations with a noise-contrastive objective \cite{gutmann2010nce,oord2018cpc} and a memory bank of negatives \cite{wu2018unsupervised}, outperforming the prior distillation methods it was compared against. RRD \cite{giakoumoglou2025relationalrepresentationdistillation} relaxes the contrastive framework by replacing instance-level discrimination with distributional alignment, minimizing KL divergence between softmax-normalized similarity distributions with separate teacher and student temperatures. While these methods capture richer geometric information than pointwise approaches, they share a common limitation that all samples and pairs are treated uniformly, spending equal gradient on regions the student has already mastered and regions where it fundamentally disagrees with the teacher.
In self-supervised contrastive learning, momentum-updated memories \cite{wu2018unsupervised,he2020moco} and hard-negative mining \cite{kalantidis2020hard,robinson2021hard} are well established; our use of RDX to select negatives can be viewed as a form of hard-negative mining in which hardness is defined by teacher--student disagreement rather than by similarity to the anchor.

\subsection{Representational Similarity and Difference}
Representational similarity methods quantify how similar two network representations are. CKA \cite{kornblith2019similarityneuralnetworkrepresentations} produces scalar similarity scores but does not identify where representations differ. Dictionary learning approaches such as sparse autoencoders \cite{cunningham2024sparse} and NMF \cite{lee2000algorithms} decompose individual representations into interpretable components but are not designed to contrast two representations against each other.
Representational Difference Explanations (RDX) \cite{kondapaneni2025representationaldifferenceexplanations} addresses this gap directly. Given two embedding matrices over the same data, RDX computes rank-normalized pairwise distances, applies a locally-biased difference function that emphasizes disagreements between nearby samples, and clusters the resulting affinity matrix to reveal groups of samples that one model considers similar but the other does not. RDX was designed as an interpretability tool which generates visual explanations of representational differences. Our work repurposes the core computational machinery of RDX, specifically the rank normalization and locally-biased difference, as an online supervision signal within a distillation loss. Rather than using RDX to produce post-hoc explanations, we use it to compute per-sample confusion weights and to select contrastive pairs, both refreshed periodically during training. To our knowledge, this is the first application of representational difference analysis as a direct component of a distillation objective.
\subsection{Curriculum Learning}
The idea of weighting training samples non-uniformly has a long history. Curriculum learning \cite{bengio2009curriculum} trains on easy examples first and gradually introduces harder ones. Self-paced learning \cite{kumar2010selfpacedlearning} automates this progression.  In the distillation context, curriculum ideas have been applied to, e.g., the distillation temperature \cite{li2023ctkd}, and other methods emphasize harder samples through temperature scaling or hard mining strategies. These approaches typically measure difficulty through output-level signals such as classification loss or prediction entropy. Our RDX-based confusion weighting instead measures per-sample difficulty from the structural disagreement between the teacher's and student's representations, and is agnostic to the architectures involved.

\section{Method}

\begin{figure}[h!]
\vskip 0.2in
\begin{center}
\centerline{\includegraphics[width=\columnwidth]{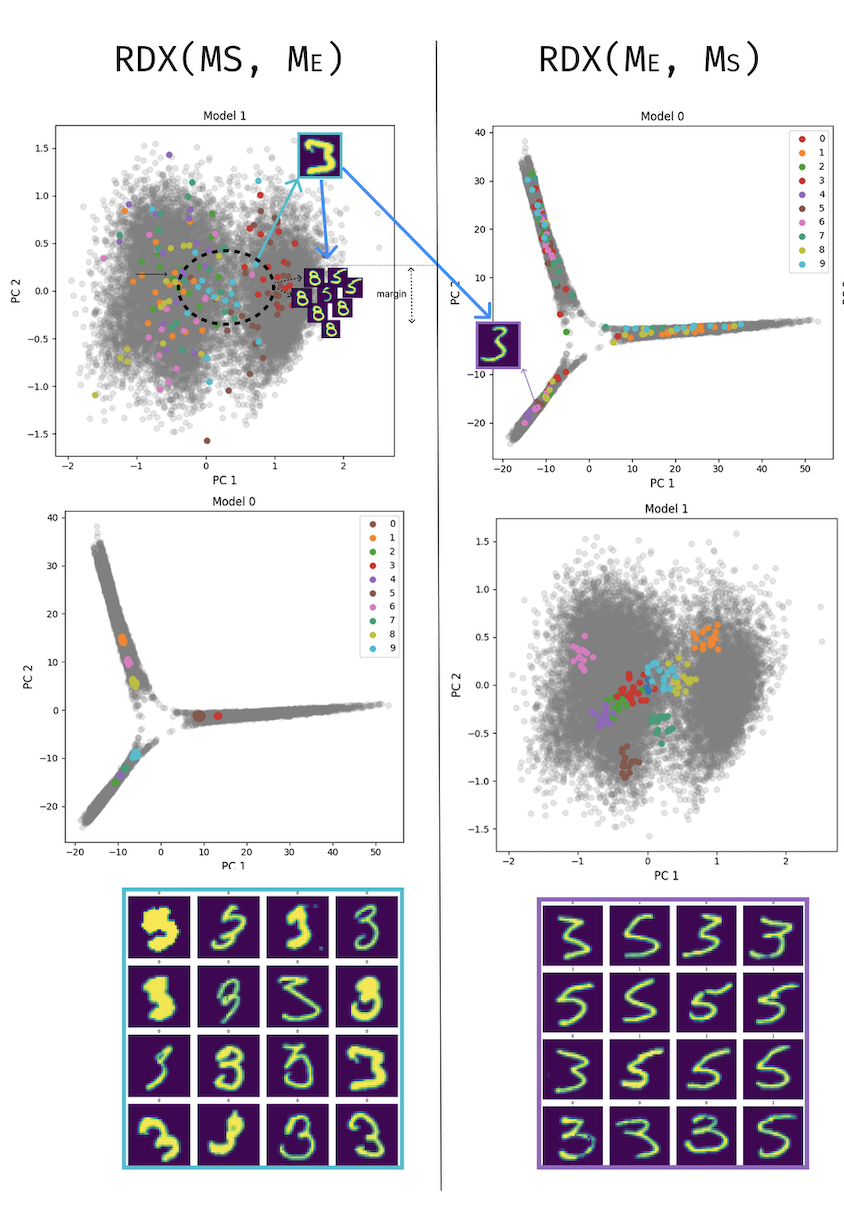}}
\caption{RDX-guided selection on MNIST \cite{lecun1998mnist} digits 3, 5, 8 with a student model $M_S$ and expert model $M_E$. We reproduce the MNIST-$\{3,5,8\}$ comparison of \citet[Sec.~4.2]{kondapaneni2025representationaldifferenceexplanations} with the official RDX code, and annotate the resulting projections to illustrate how RDX informs positive and negative selection.
\textbf{Left}: $RDX(M_S, M_E)$ reveals clusters unique to the student: samples the student groups tightly (cyan cluster, top-left PCA) that the expert separates (top-right PCA). These are the hard negatives: the student wrongly merges 3s, 5s, and 8s that the teacher distinguishes, identified by high $A_{ij}$ in the affinity matrix. \textbf{Right:} $RDX(M_E, M_S)$ reveals structure unique to the teacher: samples the expert clusters coherently that the student scatters. These inform positive selection: teacher-close pairs the student has yet to learn. Bottom grids show the corresponding image samples for each direction.}

\label{fig:triplet_selection}
\end{center}
\vskip -0.2in
\end{figure}

We propose Representational Difference Distillation (RDD), a knowledge distillation framework that uses representational difference analysis to focus the distillation loss where the teacher and student structurally disagree.

\subsection{Identifying Representational Misalignment}
\label{sec:rdx_diagnostic}

We periodically compute RDX statistics over the \emph{entire training set} to identify where the student and teacher representations disagree, following \citet{kondapaneni2025representationaldifferenceexplanations}. Let $N$ be the number of training samples and $\mathbf{F}^T \in \mathbb{R}^{N \times d_T}$, $\mathbf{F}^S \in \mathbb{R}^{N \times d_S}$ the penultimate-layer features for teacher and student. We compute RDX affinities (row-at-a-time, optionally using anchor points) to summarize mismatch and refresh these scores every $R$ epochs to track the student’s evolving representation.

We compute pairwise Euclidean distance matrices $\mathbf{D}^T, \mathbf{D}^S \in \mathbb{R}^{N \times N}$ and convert them to rank matrices: $r^T_{ij}$ is the rank of sample $j$ among all neighbors of $i$ in teacher space, and $r^S_{ij}$ the corresponding rank in student space. This rank normalization makes the comparison invariant to the scale of each feature space.

We then compute a locally-biased difference that amplifies disagreements between nearby samples while suppressing those among distant ones:
\begin{equation}
\label{eq:diff}
\text{diff}_{ij} = \tanh\!\left(\gamma \cdot \frac{r^T_{ij} - r^S_{ij}}{\min(r^S_{ij},\, r^T_{ij}) + 1}\right)
\end{equation}
where $\gamma$ controls the saturation rate. Positive values indicate the student \textbf{over-groups} $i$ and $j$ (ranks $j$ closer to $i$ than the teacher does); negative values indicate an \textbf{alignment gap}, a teacher neighbor the student has yet to bring close; values near zero indicate agreement. Because ranks are not symmetric ($r_{ij} \neq r_{ji}$ in general), neither is $\text{diff}$.

To capture both directions of mismatch, we convert this difference into two directional affinities with sharpness $\beta$:
\begin{equation}
\label{eq:affinity}
A^{01}_{ij} = \exp\!\big({-}\beta \cdot \text{diff}_{ij}\big), \qquad
A^{10}_{ij} = \exp\!\big({+}\beta \cdot \text{diff}_{ij}\big).
\end{equation}
$A^{01}$ is large for \emph{teacher-unique} pairs (close for the teacher, far for the student), and $A^{10}$ is large for \emph{student-unique} pairs (close for the student, far for the teacher). We set $A^{01}_{ii} = A^{10}_{ii} = 0$ and define a per-sample confusion score:
\begin{equation}
\label{eq:confusion}
c_i = \frac{1}{N - 1} \sum_{j \neq i} \left(A^{01}_{ij} + A^{10}_{ij}\right)
\end{equation}
Since $A^{01}_{ij} + A^{10}_{ij} = 2\cosh(\beta\,\text{diff}_{ij}) \geq 2$, with equality only when the two rankings agree, a high $c_i$ indicates that sample $i$ lies in a region of stronger teacher--student disagreement in either direction. For efficiency, all quantities are computed one block of rows at a time without materializing the full $N \times N$ matrices, and the columns $j$ can optionally be restricted to a random subset of anchor samples.

\subsection{Contrastive Supervision}
\label{rdd_loss}
Having identified \emph{where} the student misaligns with the teacher via the RDX scores, we now describe \emph{how} we correct it. We use a multi-negative contrastive objective with a memory bank (as in \citet{tian2020contrastive}), but we make two RDX-guided changes: (1) \textbf{selection}—positives and negatives are chosen using RDX affinities, and (2) \textbf{weighting}—each sample’s contrastive loss is scaled by its RDX confusion score. These two mechanisms focus gradients on regions of representational disagreement. \\\\
\textbf{Projection}
Following \citet{tian2020contrastive}, we project both teacher and student features into a shared $d$-dimensional space using linear projections $g_S : \mathbb{R}^{d_S} \to \mathbb{R}^{d}$ and $g_T : \mathbb{R}^{d_T} \to \mathbb{R}^{d}$, followed by L2 normalization:
\begin{equation}
\label{eq:projection}
\mathbf{z}^S_i = \frac{g_S(\mathbf{f}^S_i)}{\|g_S(\mathbf{f}^S_i)\|_2}, \quad \mathbf{z}^T_i = \frac{g_T(\mathbf{f}^T_i)}{\|g_T(\mathbf{f}^T_i)\|_2}
\end{equation}
This places all features on the unit hypersphere and enables a stable contrastive objective across teacher--student pairs. The student projection $g_S$ is learned. Because RDD only uses the student-anchored term of the CRD objective (student embedding contrasted against teacher memory entries), $g_T$ receives no gradient and remains a fixed random linear map.

\paragraph{Memory bank and RDX-guided sampling.}
A mini-batch provides limited negatives, so we maintain a memory bank $\mathbf{M} \in \mathbb{R}^{N \times d}$ of L2-normalized projected teacher features for all $N$ training samples \cite{wu2018unsupervised,tian2020contrastive}. RDX guides \emph{which} memory entries serve as positives and negatives. For each anchor $i$, we select the positive as the strongest teacher-unique neighbor:
\begin{equation}
\label{eq:pos_sampling}
p_i = \argmax_{j \neq i}\; A^{01}_{ij}
\end{equation}
so that the student is pulled toward a sample the teacher places close to $i$ but the student does not. Unlike CRD, whose positive is the teacher embedding of the anchor itself, the RDD positive is therefore a different sample. Negatives are drawn uniformly from a pool of the $P$ strongest student-unique neighbors:
\begin{equation}
\label{eq:neg_pool}
\mathcal{N}_i = \operatorname{Top}\text{-}P_{j \neq i}\; A^{10}_{ij}, \quad n_{i,\ell} \sim \text{Unif}(\mathcal{N}_i)
\end{equation}
where $\ell = 1,\dots,k$ indexes the $k$ negatives used in the contrastive loss (sampling with replacement when $k > P$). Given the student embedding $\mathbf{z}^S_i$, the positive $\mathbf{M}_{p_i}$ and the negatives $\{\mathbf{M}_{n_{i,\ell}}\}_{\ell=1}^{k}$, we compute the per-anchor noise-contrastive loss $\mathcal{L}^{\text{NCE}}_i$ of \citet{tian2020contrastive} with temperature $\tau$. The bank is updated every batch via exponential moving average:
\begin{equation}
\label{eq:ema}
\mathbf{M}_i \leftarrow \operatorname{normalize}\!\big(\mu \cdot \mathbf{M}_i + (1 - \mu) \cdot \mathbf{z}^T_i\big)
\end{equation}
for each sample $i$ in the current batch, with momentum $\mu$. Since both the teacher backbone and $g_T$ are fixed, the update averages each entry over the data augmentations seen during training.

\paragraph{Combined loss (RDX weighting).}
We additionally weight each sample’s contrastive loss by its normalized RDX confusion score:
\begin{equation}
\label{eq:loss_rdd}
\mathcal{L}_{\text{RDD}} = \frac{1}{B} \sum_{i=1}^{B} \frac{c_i}{\bar{c}} \cdot \mathcal{L}^{\text{NCE}}_i
\end{equation}
where $\bar{c} = \frac{1}{B}\sum_i c_i$ normalizes weights to unit mean. This focuses the distillation signal on samples where the teacher and student representations disagree most.

\subsection{Training Objective}
\label{sec:overall_objective}
The student is trained with three loss terms:
\begin{equation}
\label{eq:total_loss}
\mathcal{L} = \lambda_{\text{CE}} \cdot \mathcal{L}_{\text{CE}} + \lambda_{\text{KD}} \cdot \mathcal{L}_{\text{KD}} + \lambda_{\text{RDD}} \cdot \mathcal{L}_{\text{RDD}}
\end{equation}
where $\mathcal{L}_{\text{CE}}$ is the standard cross-entropy with ground-truth labels, $\mathcal{L}_{\text{KD}}$ is the KL divergence between temperature-scaled teacher and student logits~\cite{hinton2015distillingknowledgeneuralnetwork}, and $\mathcal{L}_{\text{RDD}}$ is our proposed loss. Setting $\lambda_{\text{KD}}=0$ recovers a pure feature-level method, which is the configuration we evaluate in \cref{sec:experiments}.

\begin{algorithm}[ht]
\caption{One training step of RDD}
\label{alg:rdd}
\begin{algorithmic}[1]
\REQUIRE Mini-batch $\{(\mathbf{x}_i, y_i, \text{idx}_i)\}_{i=1}^B$, student $f^S$, teacher $f^T$, projections $g_S, g_T$, memory bank $\mathbf{M}$, RDX tables $A^{01}, A^{10}$ and weights $\{c_i\}$ from the last refresh
\STATE Extract features $\mathbf{F}^S, \mathbf{F}^T$ and logits from $f^S, f^T$ 
\STATE Project and normalize $\mathbf{z}^S_i$, $\mathbf{z}^T_i$ \hfill \COMMENT{Eq.~\ref{eq:projection}}
\STATE Look up positive $p_i = \argmax_j A^{01}_{ij}$ and sample negatives $n_{i,\ell} \sim \text{Unif}(\operatorname{Top}\text{-}P(A^{10}_{i\cdot}))$ \hfill \COMMENT{Eqs.~\ref{eq:pos_sampling},~\ref{eq:neg_pool}}
\STATE Compute contrastive loss $\mathcal{L}^{\text{NCE}}_i$ against memory bank entries $\mathbf{M}_{p_i}$, $\mathbf{M}_{n_{i,\ell}}$
\STATE $\mathcal{L} \leftarrow \lambda_{\text{CE}} \mathcal{L}_{\text{CE}} + \lambda_{\text{KD}} \mathcal{L}_{\text{KD}} + \lambda_{\text{RDD}} \mathcal{L}_{\text{RDD}}$, where $\mathcal{L}_{\text{RDD}}=\frac{1}{B}\sum_i \frac{c_i}{\bar{c}}\mathcal{L}^{\text{NCE}}_i$ \hfill \COMMENT{Eqs.~\ref{eq:loss_rdd},~\ref{eq:total_loss}}
\STATE Backpropagate $\mathcal{L}$, step optimizer
\STATE Update memory bank: $\mathbf{M}_{\text{idx}_i} \leftarrow \text{normalize}(\mu \cdot \mathbf{M}_{\text{idx}_i} + (1-\mu) \cdot \mathbf{z}^T_i)$ \hfill \COMMENT{Eq.~\ref{eq:ema}}
\end{algorithmic}
\end{algorithm}

\section{Experiments}
\label{sec:experiments}

\subsection{Setup}

\paragraph{Dataset.}
We evaluate on CIFAR-100~\cite{krizhevsky2009learning}, which contains 50,000 training and 10,000 test images across 100 classes.

\paragraph{Teacher--student pair.}
We use a same-family pair, resnet32x4$\to$resnet8x4 \cite{he2016resnet}, following the evaluation protocol and the pretrained teacher of \citet{tian2020contrastive}.

\paragraph{Baselines.}
We compare against KD~\citep{hinton2015distillingknowledgeneuralnetwork}, RKD~\citep{park2019relationalknowledgedistillation}, and CRD~\citep{tian2020contrastive}, using the public implementation and default hyperparameters of \citet{tian2020contrastive}: KD uses $\lambda_{\text{CE}}{=}0.1$, $\lambda_{\text{KD}}{=}0.9$ and temperature 4; CRD uses $\lambda_{\text{CE}}{=}1$ and a CRD weight of $0.8$ without the KD term; RKD uses $\lambda_{\text{CE}}{=}1$ and an RKD weight of $1$. All methods use the same training recipe: SGD with momentum 0.9, weight decay $5{\times}10^{-4}$, initial learning rate 0.05 decayed by $0.1$ at epochs 150, 180, and 210, for 240 total epochs with batch size 64.

\paragraph{RDD hyperparameters.}
RDD supervision uses $\lambda_{\text{CE}}{=}1$, $\lambda_{\text{KD}}{=}0$ and $\lambda_{\text{RDD}}{=}0.8$, i.e., the same loss weights as CRD, so that the two methods differ only in the RDX-guided selection and weighting. The contrastive head follows CRD: projection dimension $d{=}128$, temperature $\tau{=}0.07$, $k{=}16{,}384$ negatives per anchor, and memory momentum $\mu{=}0.5$. Negatives are drawn from a pool of $P{=}1{,}024$ student-unique neighbors.
For the RDX diagnostic we use $\gamma{=}0.1$ and $\beta{=}0.8$, and refresh the RDX tables every $R{=}20$ epochs starting from the first epoch.
The curriculum and sampling variants keep the KD objective ($\lambda_{\text{CE}}{=}0.1$, $\lambda_{\text{KD}}{=}0.9$) and use RDX only to choose which training samples are presented; the curriculum starts from the easiest $\phi_0{=}30\%$ of samples and ramps linearly to the full set over $E_{\text{pace}}{=}100$ epochs.

\paragraph{Number of runs.}
Due to compute constraints, each configuration in this paper was trained once (a single seed), and we report the final-epoch test accuracy. Differences of a few tenths of a percent are within typical seed-to-seed variation on this benchmark and should not be read as significant.

\begin{figure}[h!]
\vskip 0.2in
\begin{center}
\centerline{\includegraphics[width=\columnwidth]{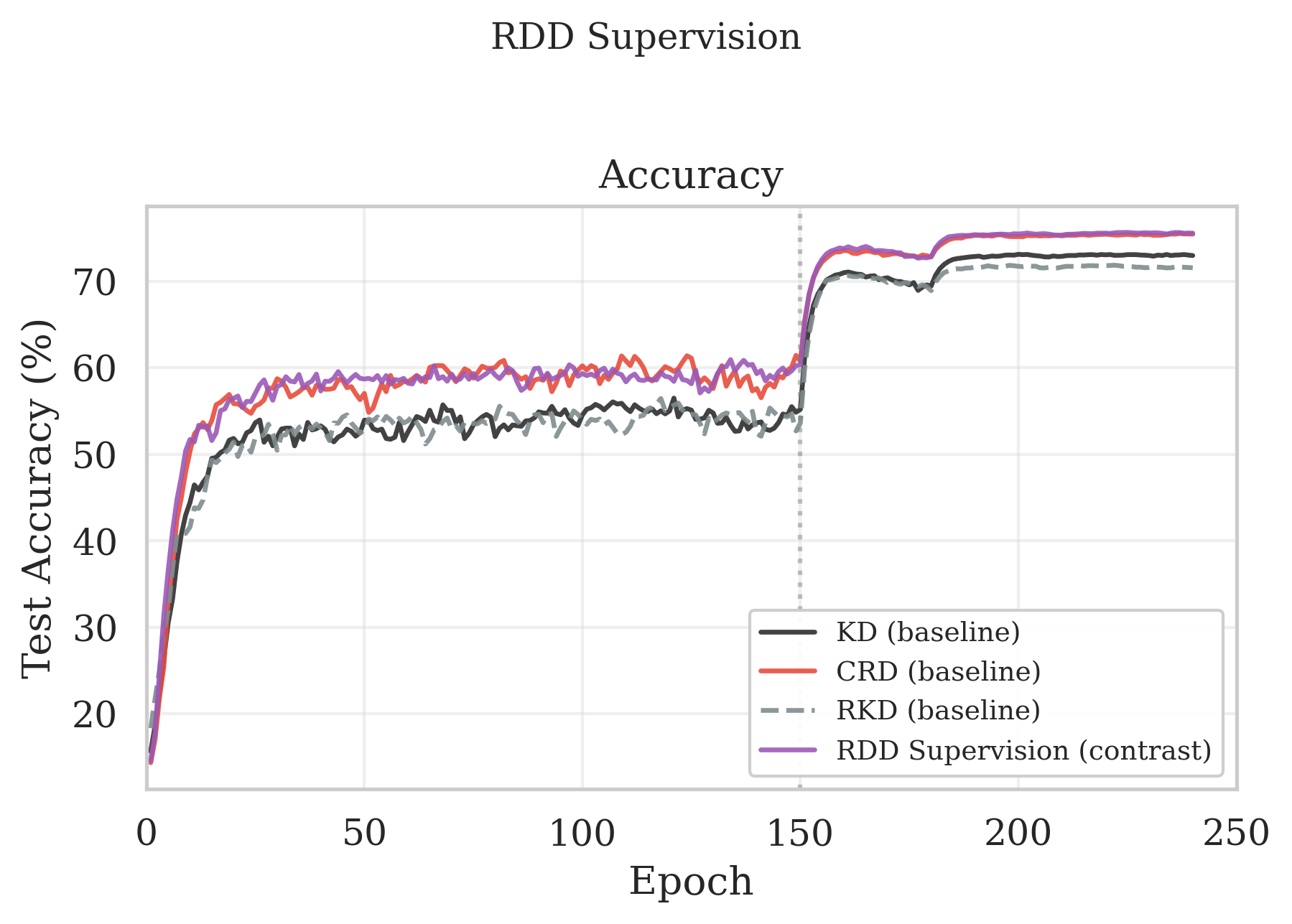}}
\caption{CIFAR-100 resnet32x4$\rightarrow$resnet8x4 test accuracy. RDD supervision uses multi-negative contrastive distillation weighted by per-sample RDX affinity to emphasize teacher--student representational disagreement. Curves shown for KD, CRD, RKD, and RDD supervision.}

\label{fig:rdd_supervision}
\end{center}
\vskip -0.2in
\end{figure}

\subsection{Results}

\paragraph{RDD supervision matches CRD and outperforms KD and RKD.}
RDD supervision reaches 75.58\% test accuracy, compared with CRD (75.45\%), KD (72.93\%), and RKD (71.55\%) (\cref{results1-table}). The 0.13-point margin over CRD comes from single runs and is within typical seed-to-seed variation, so we view the two methods as on par; the gains over KD and RKD are larger. Because RDD and CRD share the same loss weights, projection heads and memory bank, this comparison isolates the effect of RDX-guided selection and weighting. A possible explanation for the late-training behavior is the asymmetric use of the two RDX affinity matrices: while CRD samples negatives uniformly, RDD draws them from the memory entries that $A^{10}$ identifies as wrongly merged by the student. The test accuracy curves (Figure~\ref{fig:rdd_supervision}) are consistent with this interpretation: RDD supervision and CRD follow nearly identical trajectories for the first 150 epochs, and the small gap appears only after the learning-rate decay, when the remaining errors may be more localized. Multi-seed experiments and the ablations described in \cref{sec:future} are needed to confirm this.
\begin{table}[ht]
\caption{Final-epoch test accuracy (\%) for resnet32x4$\rightarrow$resnet8x4 on CIFAR-100, single run per method. The teacher (resnet32x4) reaches 79.42\% \citep{tian2020contrastive}.}
\label{results1-table}
\vskip 0.15in
\begin{center}
\begin{small}
\begin{sc}
\begin{tabular}{lc}
\toprule
\textbf{Method} & \textbf{Test Acc.} \\
\midrule
KD  \cite{hinton2015distillingknowledgeneuralnetwork}  & 72.93 \\
CRD \cite{tian2020contrastive} & 75.45 \\
RKD \cite{park2019relationalknowledgedistillation}    & 71.55 \\
RDD Sampling (ours) & 59.49 \\
RDD Curriculum (ours) & 72.96 \\
RDD Supervision (ours) & 75.58 \\
\bottomrule
\end{tabular}
\end{sc}
\end{small}
\end{center}
\vskip -0.1in
\end{table}

\begin{figure}[h!]
\vskip 0.2in
\begin{center}
\centerline{\includegraphics[width=\columnwidth]{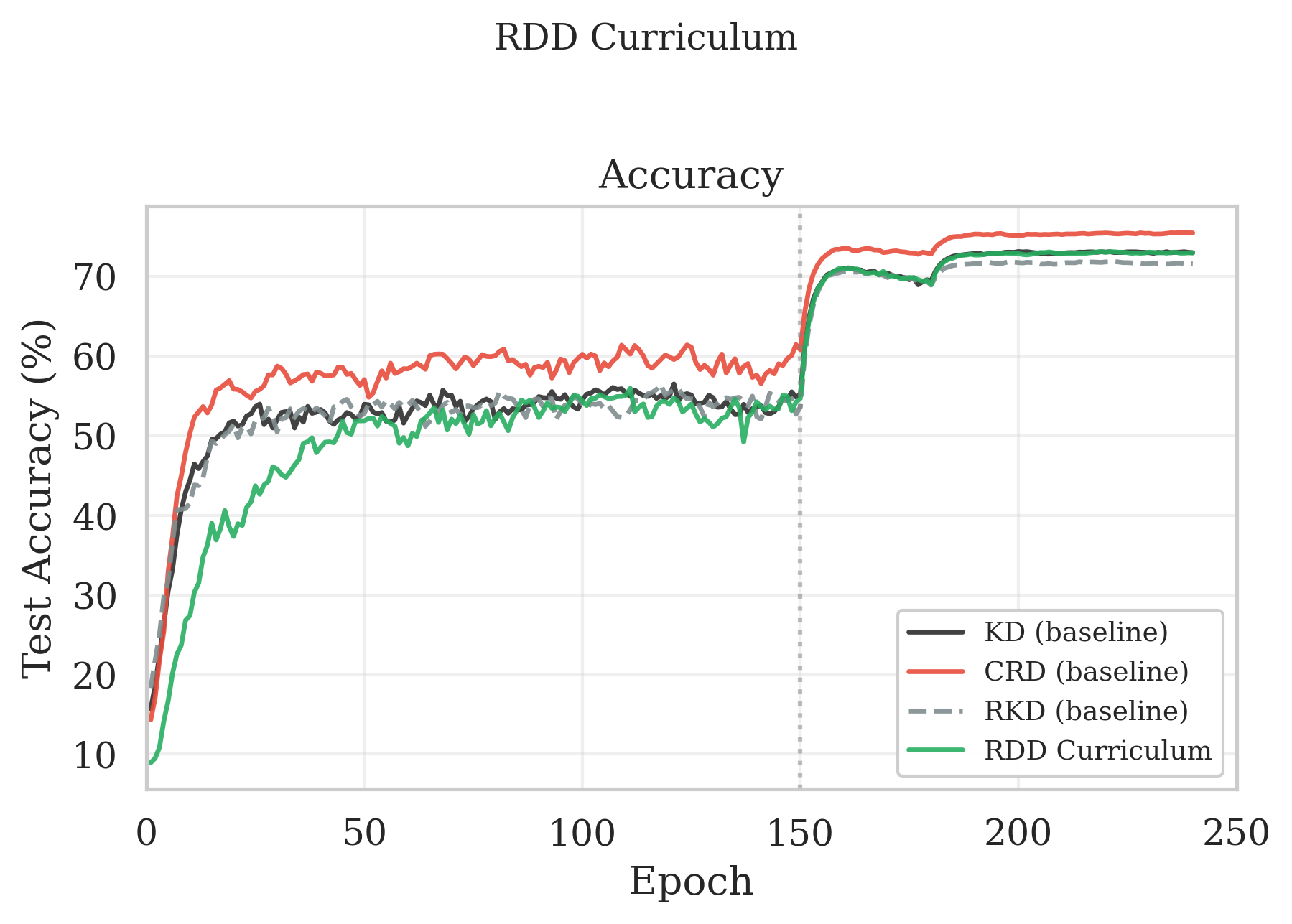}}
\caption{CIFAR-100 resnet32x4$\rightarrow$resnet8x4 test accuracy. RDD curriculum uses RDX difficulty scores to start from a fraction $\phi_0$ of the easiest samples and linearly ramps to full data over $E_{\text{pace}}$ epochs. Curves shown for KD, CRD, RKD, and RDD curriculum.}
\label{fig:rdd_curriculum}
\end{center}
\vskip -0.2in
\end{figure}

\paragraph{An RDX curriculum recovers KD performance but does not improve on it.}
RDD curriculum (72.96\%) matches KD (72.93\%), which uses the same loss, but falls short of CRD and RDD supervision. The training dynamics (Figure~\ref{fig:rdd_curriculum}) suggest why: during the initial phase when only the easiest $\phi_0 = 30\%$ of samples are available, the student learns a reasonable but incomplete representation. As harder samples are introduced, the student must reorganize its feature space to accommodate them, causing a temporary slowdown visible around epochs 60--100. By epoch 150, all samples are active and the student reaches performance comparable to KD. Since the curriculum and supervision variants also differ in their loss (KD versus contrastive), this comparison does not isolate the effect of data selection; still, it suggests that controlling \emph{which} samples the student sees is less effective than controlling \emph{how strongly} the loss reacts to each pair.

\begin{figure}[h!]
\vskip 0.2in
\begin{center}
\centerline{\includegraphics[width=\columnwidth]{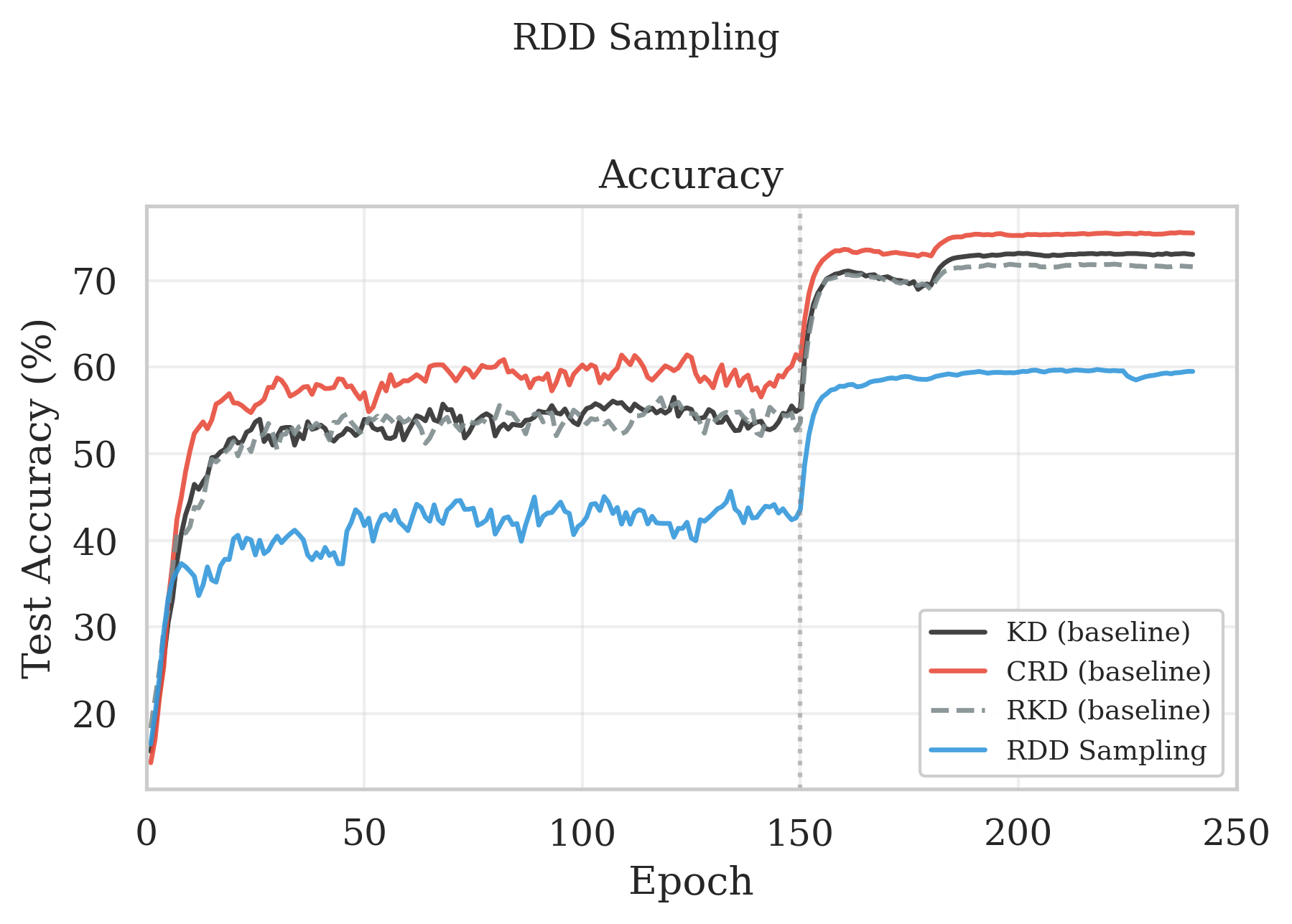}}
\caption{CIFAR-100 resnet32x4$\rightarrow$resnet8x4 test accuracy. RDD sampling trains on an RDX-ranked subset of 15\% of the training set; the subset is recomputed every $R$ epochs. Curves shown for KD, CRD, RKD, and RDD sampling.}

\label{fig:rdd_sampling}
\end{center}
\vskip -0.2in
\end{figure}

\paragraph{Naive sampling collapses without diversity.}
RDD sampling (59.49\%) performs substantially below all other methods, as shown in Figure~\ref{fig:rdd_sampling}. Without refreshing the RDX ranking, the same variant reaches only 54.53\%. This negative result is informative. By restricting training to the easiest RDX-ranked samples without a schedule to reintroduce harder ones, the student sees only the portion of the data where teacher and student already agree. It never encounters the challenging boundary cases that drive representation refinement. The student converges quickly to a solution that handles easy samples well but generalizes poorly. This failure mode is consistent with the motivation for curriculum learning~\citep{bengio2009curriculum}, where easy examples are a starting point rather than the full training signal. The contrast with RDD curriculum, which ramps to full data over $E_{\text{pace}}$ epochs, confirms that the schedule is essential.

\begin{figure}[t]
\centering
\begin{minipage}[t]{0.24\linewidth}
  \centering
  \includegraphics[width=\linewidth]{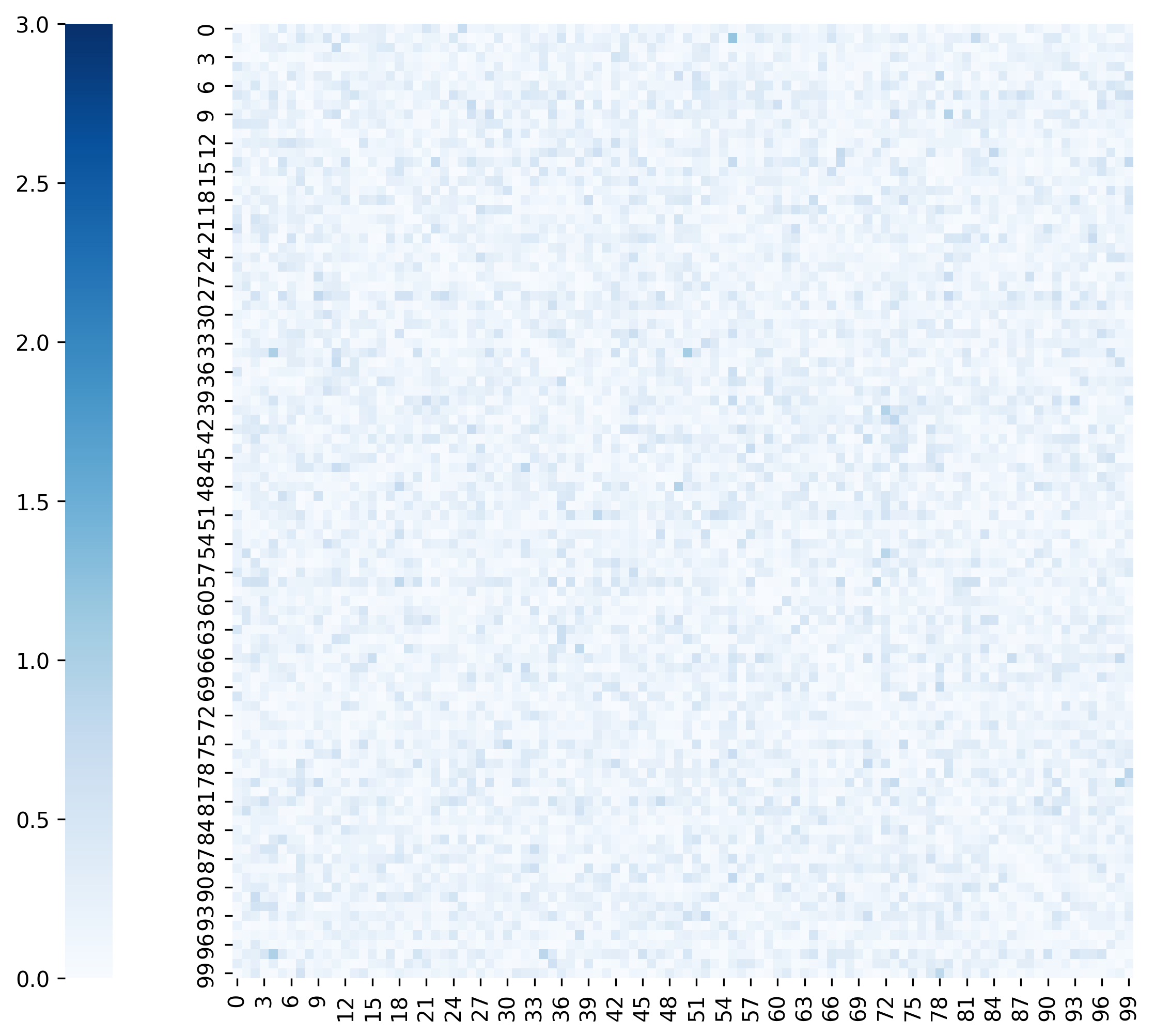}
  \\[-0.2em]\scriptsize (a) KD \\ Mean: 0.175, Max: 1.188
\end{minipage}
\begin{minipage}[t]{0.24\linewidth}
  \centering
  \includegraphics[width=\linewidth]{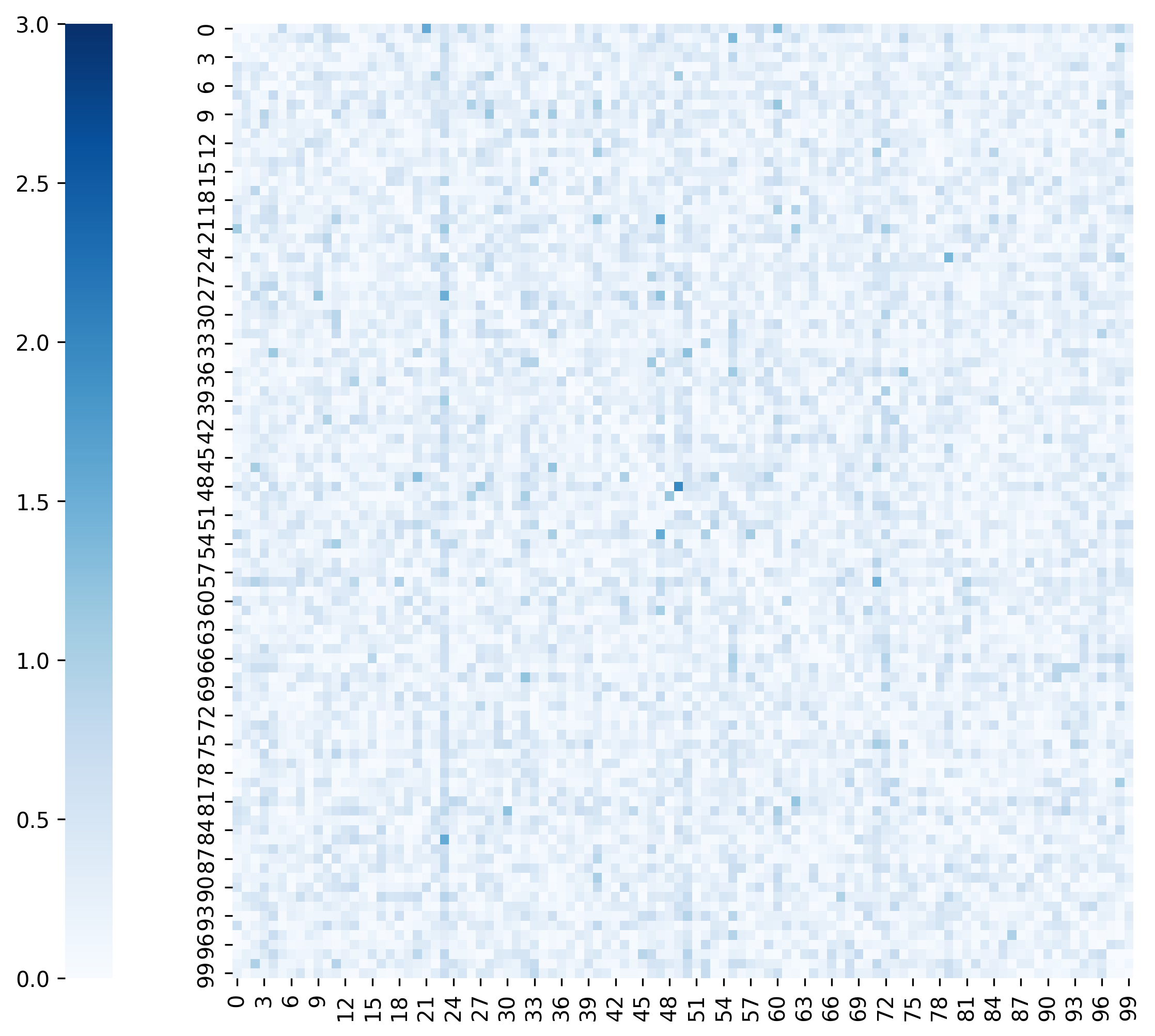}
  \\[-0.2em]\scriptsize (b) CRD \\ Mean: 0.255, Max: 1.987
\end{minipage}\hfill
\begin{minipage}[t]{0.24\linewidth}
  \centering
  \includegraphics[width=\linewidth]{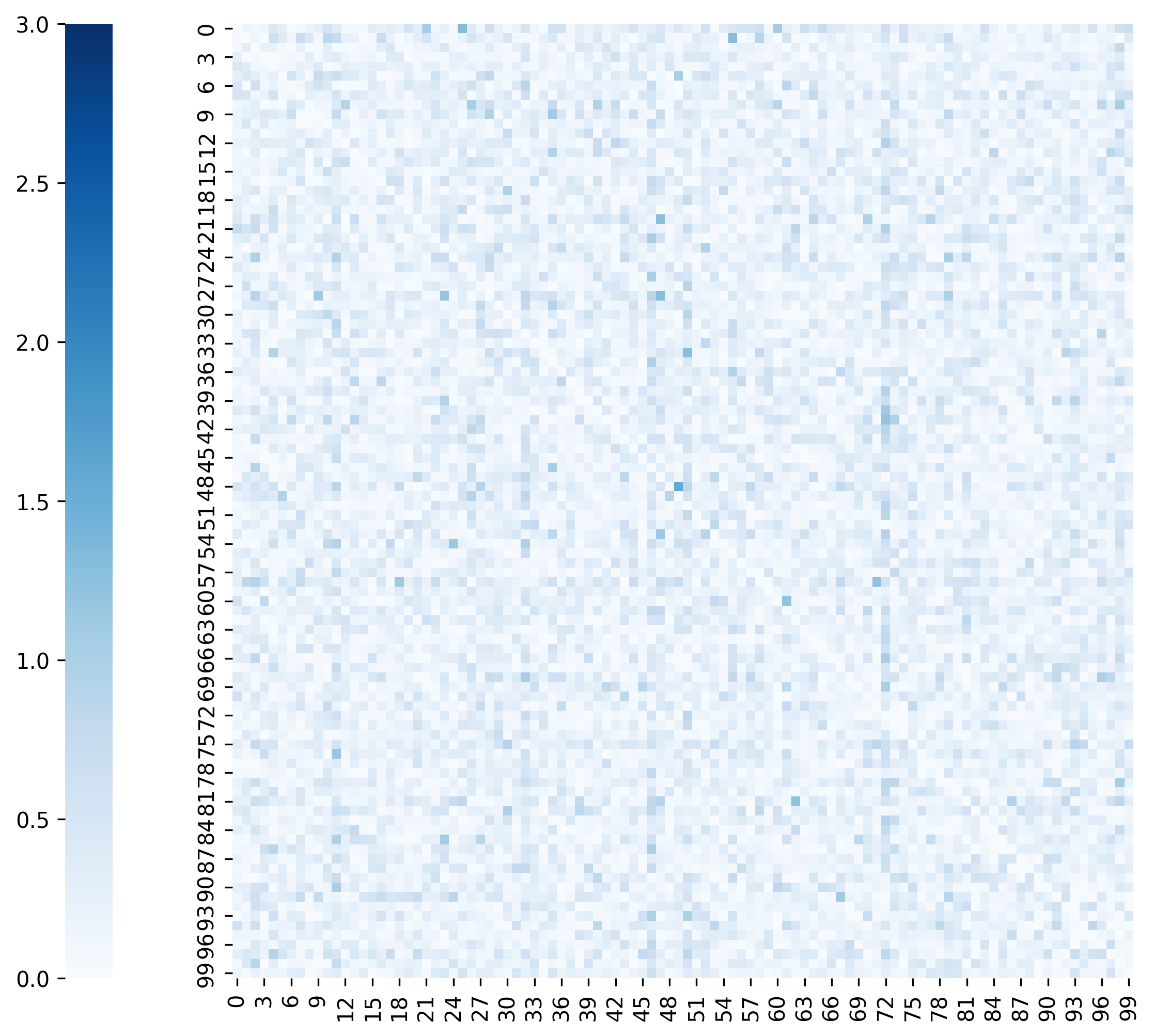}
  \\[-0.2em]\scriptsize (c) RKD \\ Mean: 0.235, Max: 1.551
\end{minipage}\hfill
\begin{minipage}[t]{0.24\linewidth}
  \centering
  \includegraphics[width=\linewidth]{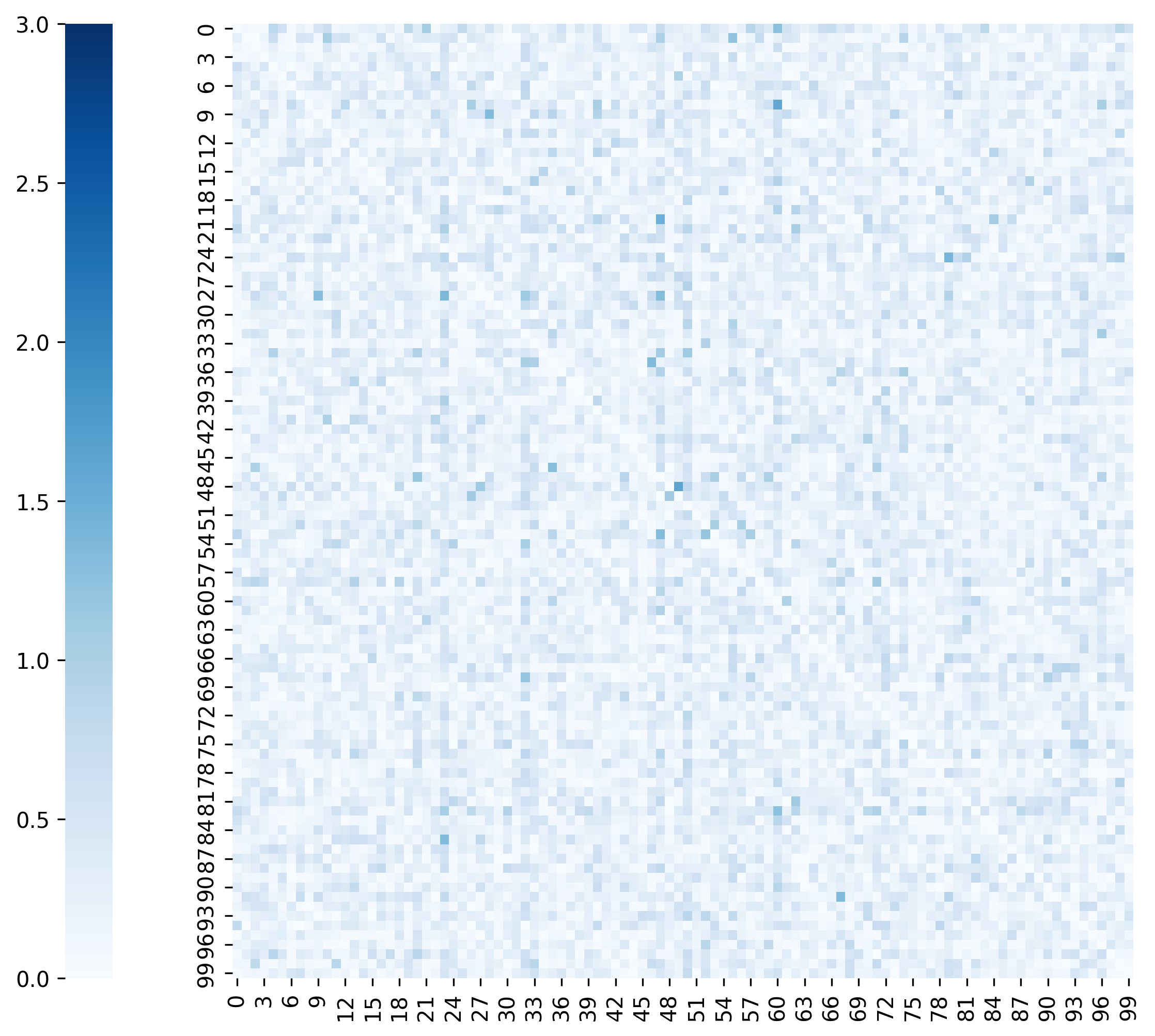}
  \\[-0.2em]\scriptsize (d) RDD (ours) \\ Mean: 0.253, Max: 1.623 
\end{minipage}\hfill
\caption{\textbf{Correlation matrix between teacher and student logits on  CIFAR-100.} Each heatmap shows the absolute difference between the class-level correlation matrices of the teacher (resnet32x4) and the student (resnet8x4) on CIFAR-100. Darker regions indicate larger discrepancy; lighter regions indicate the student has successfully captured the teacher's inter-class correlation structure. Following the visualization protocol of~\citet{tian2020contrastive}.}
\label{fig:correlation}
\end{figure}

\paragraph{Logit correlation structure.}
Figure~\ref{fig:correlation} provides a complementary view of distillation quality beyond classification accuracy, following \citet{tian2020contrastive}. The heatmaps show the absolute difference between teacher and student class-correlation matrices at the logit level. KD, which directly matches the teacher's logits, has the lowest discrepancy (mean 0.175). The feature-level methods have higher discrepancy, with RDD (mean 0.253) on par with CRD (0.255) and RKD (0.235). Thus, although the RDD loss operates on penultimate-layer features rather than logits, it preserves the teacher's inter-class correlation structure about as well as CRD, but not as well as directly matching logits.

\begin{figure}[t]
\centering
\begin{minipage}[t]{0.24\linewidth}
  \centering
  \includegraphics[width=\linewidth]{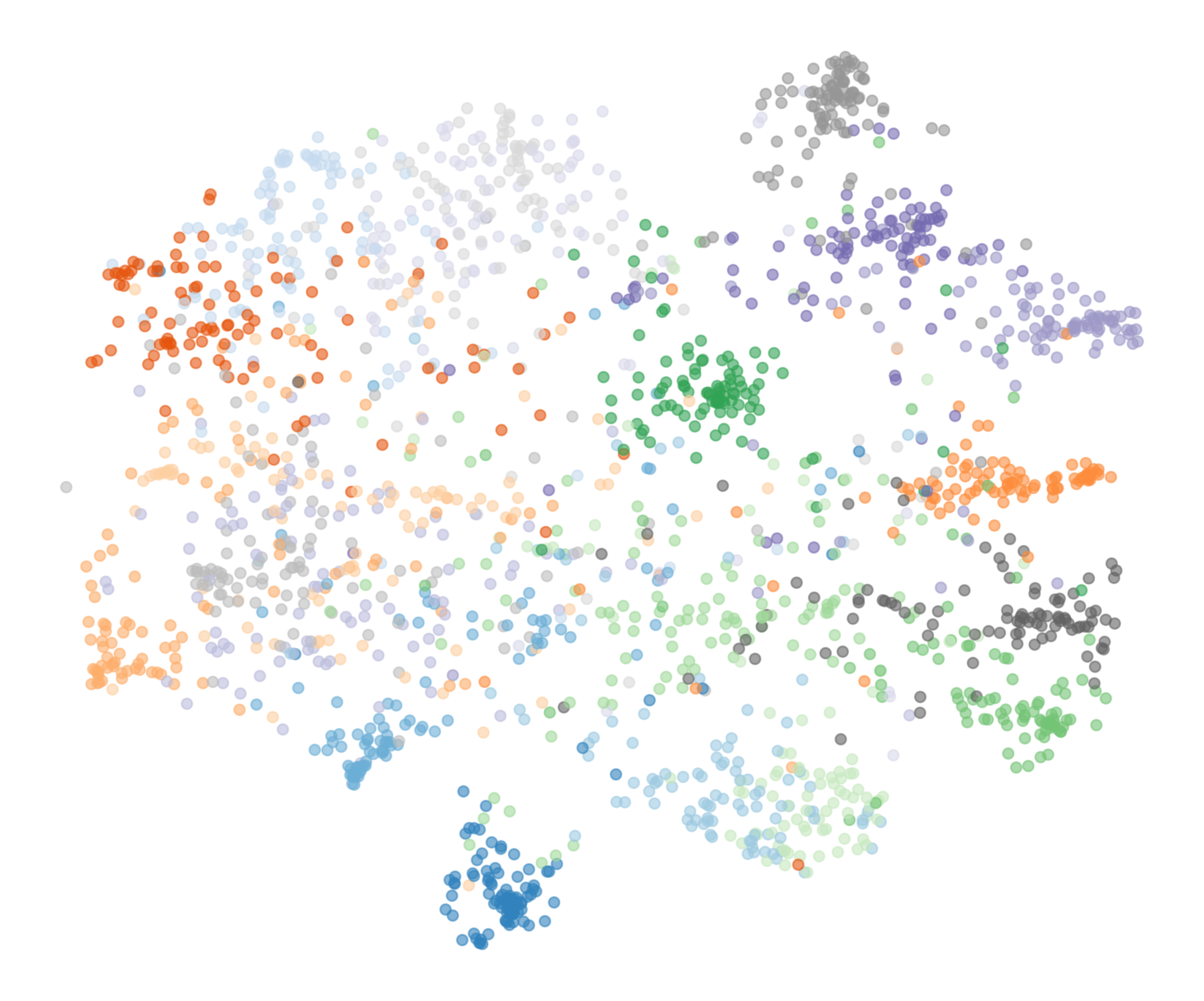}
  \\[-0.2em]\scriptsize (a) KD
\end{minipage}
\begin{minipage}[t]{0.24\linewidth}
  \centering
  \includegraphics[width=\linewidth]{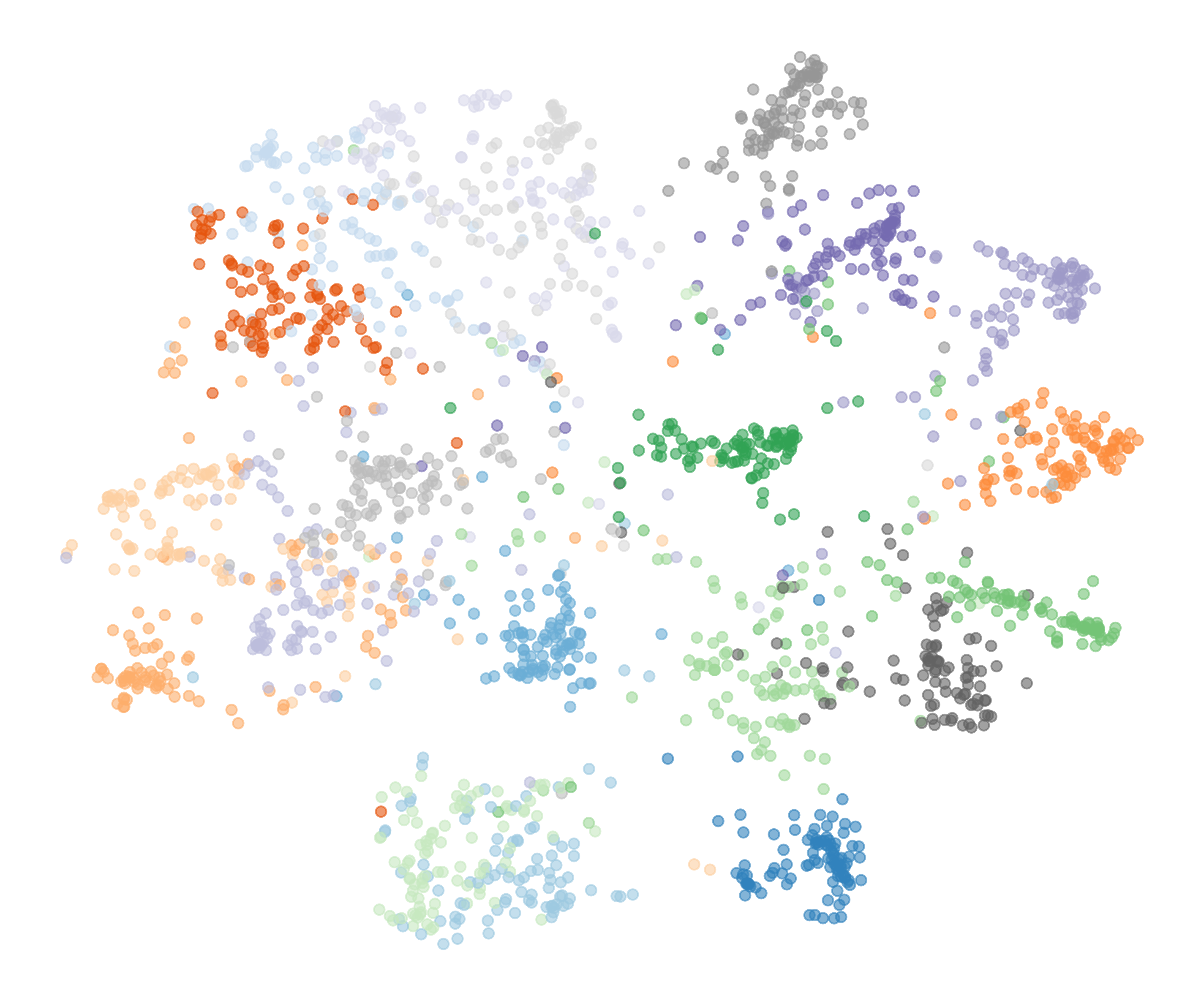}
  \\[-0.2em]\scriptsize (b) CRD 
\end{minipage}\hfill
\begin{minipage}[t]{0.24\linewidth}
  \centering
  \includegraphics[width=\linewidth]{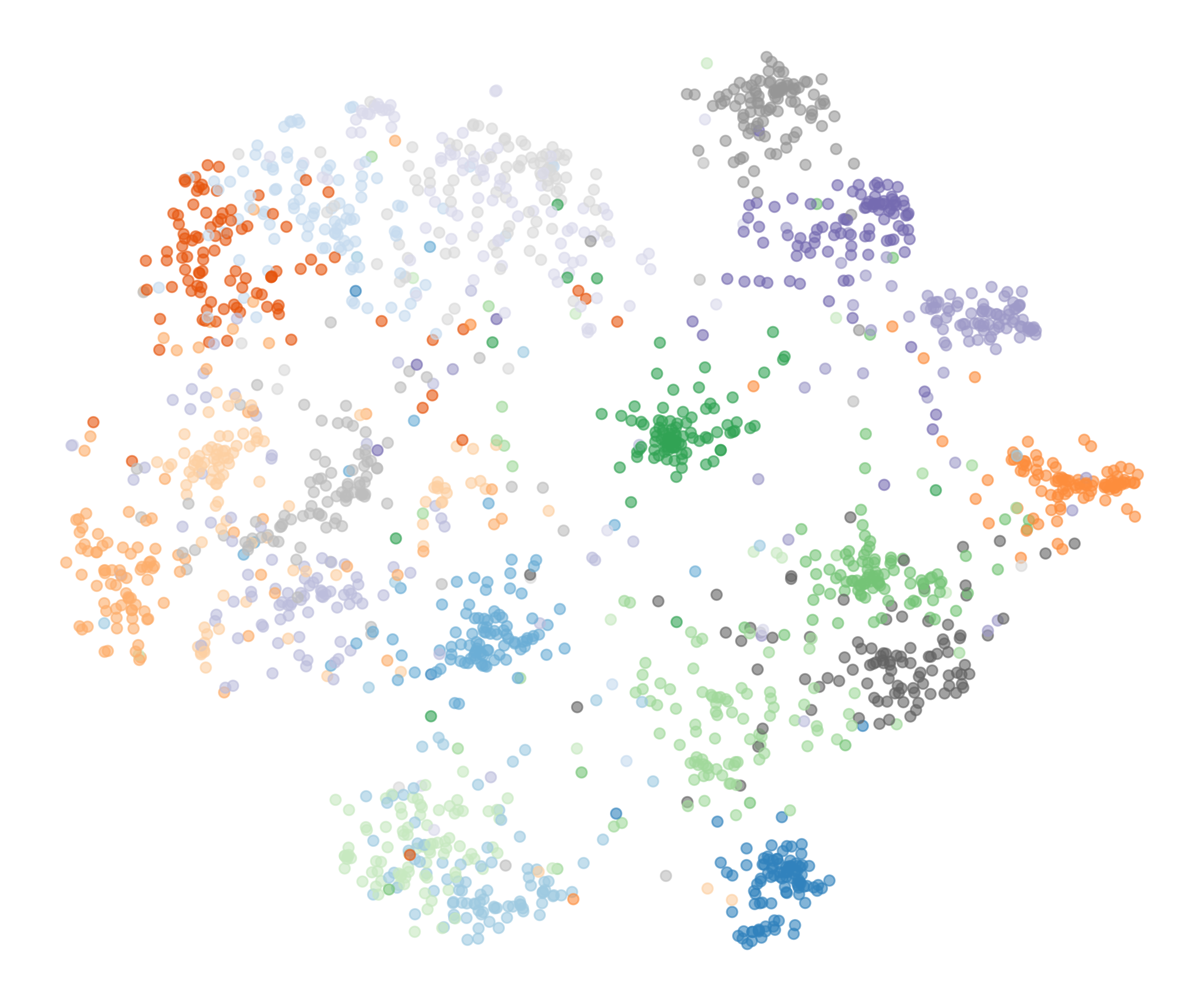}
  \\[-0.2em]\scriptsize (c) RKD
\end{minipage}\hfill
\begin{minipage}[t]{0.24\linewidth}
  \centering
  \includegraphics[width=\linewidth]{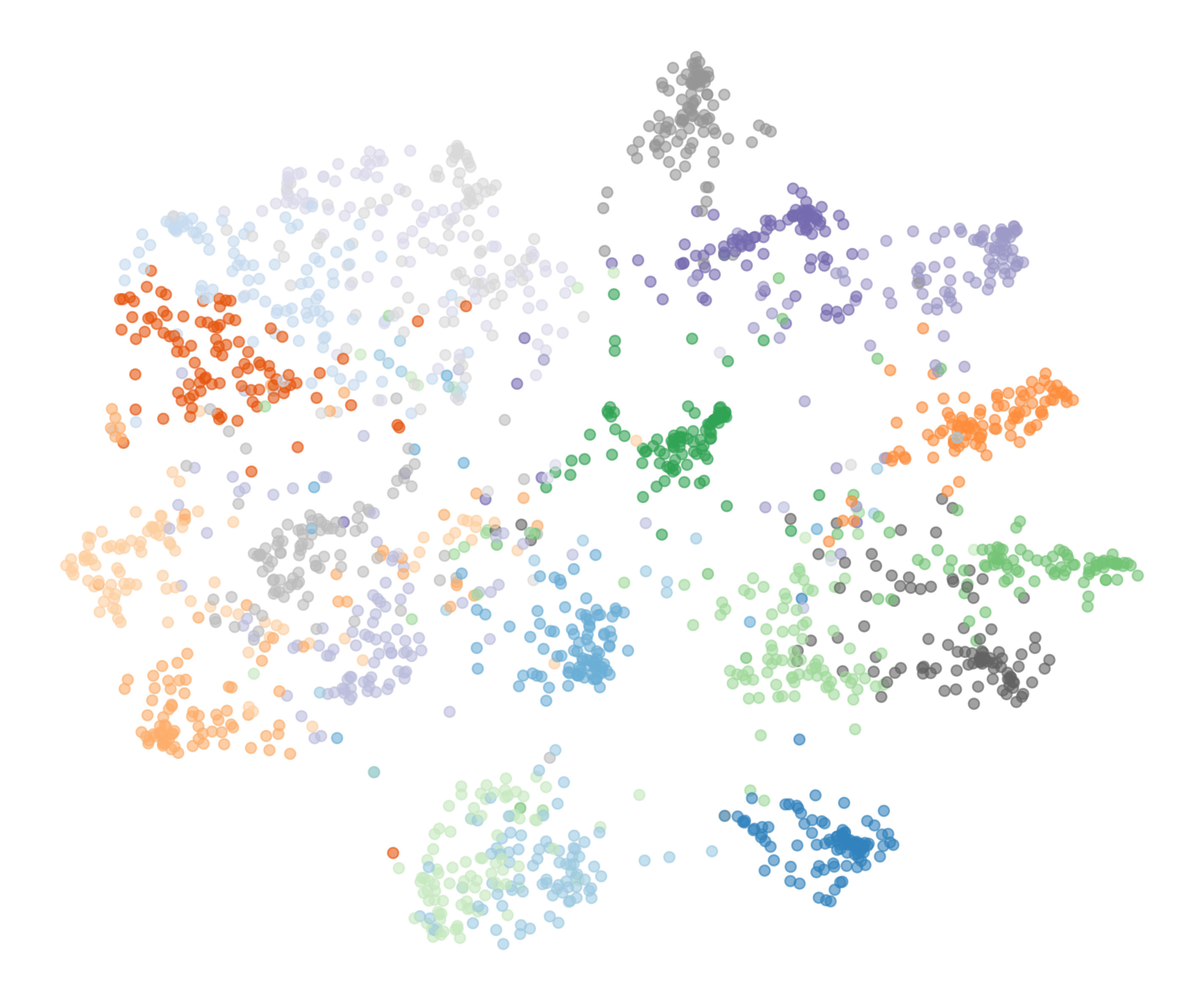}
  \\[-0.2em]\scriptsize (d) RDD (ours) 
\end{minipage}\hfill
\caption{\textbf{t-SNE visualization of student embeddings on CIFAR-100.} We plot the penultimate-layer features of the student network (resnet8x4 distilled from resnet32x4) for 20 randomly selected classes. Each color represents a distinct class. Tighter, more separated clusters indicate better representational structure transferred from the teacher. Following the visualization protocol of~\citet{tian2020contrastive}}
\label{fig:tsne}
\end{figure}

\paragraph{t-SNE embeddings.}
The t-SNE \cite{vandermaaten2008tsne} visualizations (Figure~\ref{fig:tsne}) give a qualitative view of the student representations. Both CRD and RDD (Figure~\ref{fig:tsne}b,d) produce compact intra-class clusters with clear inter-class separation. Differences between the two are subtle in these projections, and since t-SNE layouts depend on the random seed and perplexity, we do not draw quantitative conclusions from them.

\section{Discussion}

We introduced Representational Difference Distillation (RDD), which uses RDX to target teacher--student representational disagreement during distillation. On CIFAR-100, RDD supervision performs on par with CRD and above KD and RKD while keeping the overall training recipe intact, and our experiments indicate that reweighting the loss is more effective than filtering the training data. Future work includes scaling RDD to larger datasets, exploring more efficient or online RDX estimators, and evaluating distillation across more diverse teacher–student architecture pairs.

\subsection{Limitations}

\paragraph{RDX refresh staleness.}
RDX affinities are computed over the full training set but only refreshed every $R$ epochs, and held fixed between refreshes. As the student evolves, these affinities can become stale, so the supervision may lag behind the current representational geometry. More frequent refreshes reduce this lag but increase computational overhead, while less frequent refreshes trade accuracy for efficiency. An online or adaptive refresh schedule could mitigate this without constant full-dataset recomputation.

\paragraph{Statistical uncertainty.}
Each configuration was trained with a single seed, and we do not report confidence intervals. The difference between RDD supervision and CRD is smaller than typical seed-to-seed variation, so multi-seed runs are needed before claiming an improvement.

\paragraph{Computational cost.}
Each RDX refresh ranks all pairwise distances, which costs $O(N^2)$ distance computations and $O(N^2 \log N)$ for sorting over the training set (or $O(NM)$ with $M$ anchors). We did not measure the resulting wall-clock overhead relative to CRD.

\paragraph{Architectural and dataset scope.}
Our evaluation is limited to a single teacher--student pair (resnet32x4$\to$resnet8x4) on CIFAR-100. Whether RDD's gains hold for cross-architecture transfers (e.g., ResNet to MobileNet) or at ImageNet scale remains to be validated.

\paragraph{Hyperparameter sensitivity.}
RDD introduces RDX-specific parameters ($\gamma$, $\beta$) on top of the standard distillation weights. We did not conduct a sensitivity analysis. Since the positive and the negative pool depend only on the ordering of $\text{diff}_{ij}$, the sharpness $\beta$ affects only the confusion weights $c_i$: a high $\beta$ concentrates the loss on a few highly confused anchors, while $\beta \to 0$ makes the weights uniform but retains the RDX-selected pairs. How this trade-off interacts with the contrastive loss warrants further study.

\subsection{Future Directions}
\label{sec:future}
\paragraph{Ablation coverage.}
We compare three RDD variants (sampling, curriculum, supervision) but do not isolate individual components within the supervision loss. Ablations separating the contributions of $A^{01}$-guided positive selection, $A^{10}$-weighted negatives, and $c_i$ confusion weighting would clarify which mechanism drives the gains. Similarly, we do not ablate the number of negatives $k$, the negative pool size $P$, the refresh interval $R$, or the effect of combining RDD with KD versus using it standalone. These ablations are necessary to fully understand the method and guide practitioners in applying it to new settings.

Other key directions include: (1) scaling RDX to larger datasets via approximate rank computation or incrementally maintained global affinities; and (2) adaptive refresh scheduling that recomputes RDX scores based on shifts in the confusion distribution rather than at fixed intervals.

\section*{Impact Statement}
This paper presents work whose goal is to advance the field of machine learning, specifically the compression of neural networks through knowledge distillation. Smaller distilled models can reduce the energy and hardware cost of deployment. We do not foresee societal consequences specific to this work beyond those of model compression in general, including that a distilled student may inherit the biases of its teacher.

\section*{Acknowledgements}
We thank Neehar Kondapaneni and Pietro Perona for their guidance and helpful discussions. Our implementation builds on the RepDistiller codebase and pretrained teachers released by \citet{tian2020contrastive}, and on the RDX method of \citet{kondapaneni2025representationaldifferenceexplanations}.

\bibliography{example_paper}
\bibliographystyle{icml2025}

\newpage
\appendix
\onecolumn

\end{document}